\documentclass{article}

\usepackage{microtype}
\usepackage{graphicx}
\usepackage{subcaption}
\usepackage{booktabs} 

\usepackage{hyperref}

  \usepackage[preprint]{icml2026}

\usepackage{amsmath}
\usepackage{amssymb}
\usepackage{mathtools}
\usepackage{amsthm}

\usepackage{pifont}
\newcommand{\cmark}{\ding{51}} 
\newcommand{\xmark}{\ding{55}} 
\usepackage{subcaption}
\usepackage{svg}
\usepackage[table]{xcolor}
\usepackage{tcolorbox}
\usepackage{CJKutf8}
\tcbuselibrary{listings, skins, breakable}

\usepackage[capitalize,noabbrev]{cleveref}

\theoremstyle{plain}

\theoremstyle{definition}

\theoremstyle{remark}

\usepackage[textsize=tiny]{todonotes}

\newcommand{\intern}{\textsuperscript{*} Work done during an internship at Kuaishou Technology.}
\newcommand{\samename}{The fifth author is Yue Zhang (\begin{CJK}{UTF8}{gbsn}章岳\end{CJK}) and the last author is Yue Zhang (\begin{CJK}{UTF8}{gbsn}张岳\end{CJK}).}
\icmltitlerunning{Detecting RLVR Training Data}
\begin{document}

\twocolumn[
  \icmltitle{Detecting RLVR Training Data via Structural Convergence of Reasoning}



  \icmlsetsymbol{intern}{*}
  \icmlsetsymbol{corresp}{$\dagger$}

  \begin{icmlauthorlist}
    \icmlauthor{Hongbo Zhang}{zju,wlu,intern}
    \icmlauthor{Yang Yue}{ks}
    \icmlauthor{Jianhao Yan}{wlu}
    \icmlauthor{Guangsheng Bao}{wlu}
    \icmlauthor{Yue Zhang}{}
    \icmlauthor{Yue Zhang}{wlu,corresp}
  \end{icmlauthorlist}

  \icmlaffiliation{zju}{Zhejiang University}
  \icmlaffiliation{wlu}{School of Engineering, Westlake University}
  \icmlaffiliation{ks}{Kuaishou Technology}

  \icmlcorrespondingauthor{Yue Zhang}{zhangyue@westlake.edu.cn}

  \icmlkeywords{Machine Learning, ICML}

  \vskip 0.3in
]


\printAffiliationsAndNotice{\intern\ \samename}  
\newcommand{\ours}{Min-$k$NN Distance}

\begin{abstract}

Reinforcement learning with verifiable rewards (RLVR) is central to training modern reasoning models, but the undisclosed training data raises concerns about benchmark contamination. Unlike pretraining methods, which optimize models using token-level probabilities, RLVR fine-tunes models based on reward feedback from self-generated reasoning trajectories, making conventional likelihood-based detection methods less effective. We show that RLVR induces a distinctive behavioral signature: prompts encountered during RLVR training result in more rigid and similar generations, while unseen prompts retain greater diversity. We introduce \ours, a simple black-box detector that quantifies this collapse by sampling multiple completions for a given prompt and computing the average of the $k$ smallest nearest-neighbor edit distances. \ours\ requires no access to the reference model or token probabilities. Experiments across multiple RLVR-trained reasoning models show that \ours\ reliably distinguishes RL-seen examples from unseen ones and outperforms existing membership inference and RL contamination detection baselines. The project page is available at \url{https://stevenzhb.github.io/detect-rlvr-data/}.

\end{abstract}

\section{Introduction}
Recent advances in large language models (LLMs) have demonstrated remarkable improvements in reasoning performance, particularly on benchmarks with verifiable answers such as mathematics~\cite{guo2025deepseek,zeng2025simplerl,team2025kimi,he2025justrl}, coding~\cite{seed2025seed,liu2025prorl}, and symbolic problem solving~\cite{xie2025logic}. A key driver behind these gains is reinforcement learning with verifiable rewards (RLVR), in which models are trained on problems with automatically checkable outcomes, most commonly math and code tasks, and are optimized to produce correct final answers. By directly reinforcing reasoning trajectories that lead to correct solutions, RLVR enables models to refine their chain-of-thought generation beyond supervised fine-tuning. As a result, RLVR has become a cornerstone of modern post-training pipelines and now underpins many state-of-the-art reasoning models.

Despite impressive benchmark results, a growing body of evidence suggests that high reported reasoning performance does not always translate to robust generalization. When evaluation shifts to temporally newer, less exposed, or expert-level problems, model performance often drops substantially~\cite{petrov2025proof,glazer2024frontiermath,mahdavi2025leveraging}. In practice, models that solve standard benchmark questions reliably may fail on newly released problems, alternative formulations of known tasks, or slightly more complex variants that require the same underlying reasoning. These patterns raise concerns that some gains reflect over-specialization during post-training, rather than consistently robust reasoning ability. This issue is exacerbated by current open-source practices around RLVR. Many recent releases provide only RLVR-tuned models, without access to the base checkpoints or RL training data, making it difficult to assess whether benchmark problems or close paraphrases were encountered during training. As a result, risks of benchmark contamination and inflated generalization claims increase, while practitioners lack reliable tools to detect overlap with prior RLVR exposure. These challenges highlight the need for methods that can identify whether a specific example has appeared during RLVR training.

\begin{figure*}[t]
    \centering
    \includegraphics[width=\linewidth]{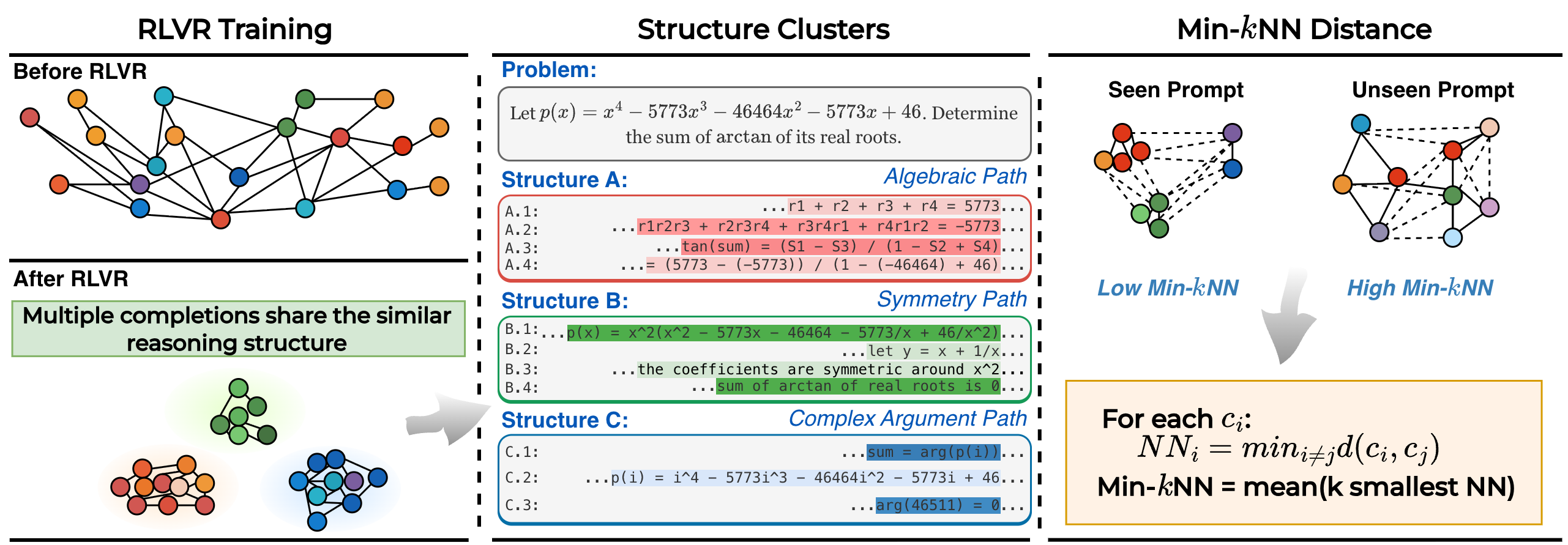}
    \caption{
    We study how RLVR induces structural convergence in reasoning trajectories. \textbf{Left}: RLVR compresses diverse reasoning paths into shared structural modes. \textbf{Middle}: multiple completions cluster into few reasoning structures. \textbf{Right}: Min-$k$NN Distance quantifies this collapse via nearest-neighbor edit distances, yielding low values for seen prompts and high values for unseen prompts.
    }
    \label{fig:overview}
\end{figure*}
Prior work has shown that detecting training data exposure during pretraining or supervised fine-tuning is often feasible, as memorization under likelihood-based objectives leaves strong statistical traces detectable even in black-box settings~\cite{shi2023detecting,zhang2024min,xie2024recall,dong2024generalization}. However, detecting exposure during RLVR training presents a different challenge. Unlike likelihood-based training, RLVR optimizes models through reward feedback on self-generated reasoning trajectories, which makes conventional token-level or likelihood-based signals ineffective. As demonstrated in our experiments (Section~\ref{sec:exp}), traditional detection methods, which rely on perplexity and token-level statistics, fail to capture the changes in reasoning patterns induced by RLVR.
To understand how RLVR reshapes model behavior, we analyze the evolution of generation diversity during training. As shown in Fig.~\ref{fig:overview}, RLVR induces a systematic convergence in reasoning trajectories: prompts seen during RL training yield increasingly similar generations, while unseen prompts retain high variability. This collapse concentrates on symbolic and algebraic reasoning components and compresses outputs into a small number of recurring structural modes rather than a single derivation. These stable structural patterns form a distinctive behavioral signature of RLVR exposure.

Building on this observation, we introduce \ours, a simple black-box method for detecting whether an example has appeared during RLVR training. As illustrated in Fig.~\ref{fig:overview}, given a prompt example, \ours\ samples multiple completions from the RLVR-tuned model and quantifies their structural diversity by computing edit distances among generated outputs. Since RLVR induces a convergence in reasoning diversity for seen examples, prompts that appeared during RLVR training yield consistently smaller \ours\ values than unseen prompts. The method requires only sampling access to the RLVR-tuned model and does not rely on token log probabilities or any reference models. Extensive experiments across multiple reasoning models and RLVR setups demonstrate that \ours\ reliably distinguishes RL-seen examples from unseen ones.

Our contributions are as follows:
\begin{itemize}
    \item We provide the first systematic analysis of how RLVR reshapes reasoning behavior, revealing a convergence in the structural space of reasoning trajectories, particularly in symbolic reasoning components.
    \item Based on these findings, we introduce \ours, a simple, black-box method for detecting RLVR exposure without requiring access to the token log probabilities.
    \item We conduct extensive experiments across diverse reasoning models, RL algorithms, and training setups, showing that \ours\ reliably distinguishes RL-seen examples from unseen ones, even under challenging conditions like paraphrasing, distillation, and varying decoding configurations.
\end{itemize}

\section{Preliminary}
\begin{figure*}[t]
    \centering
    \begin{subfigure}[t]{0.32\linewidth}
        \centering
        \includegraphics[width=\linewidth]{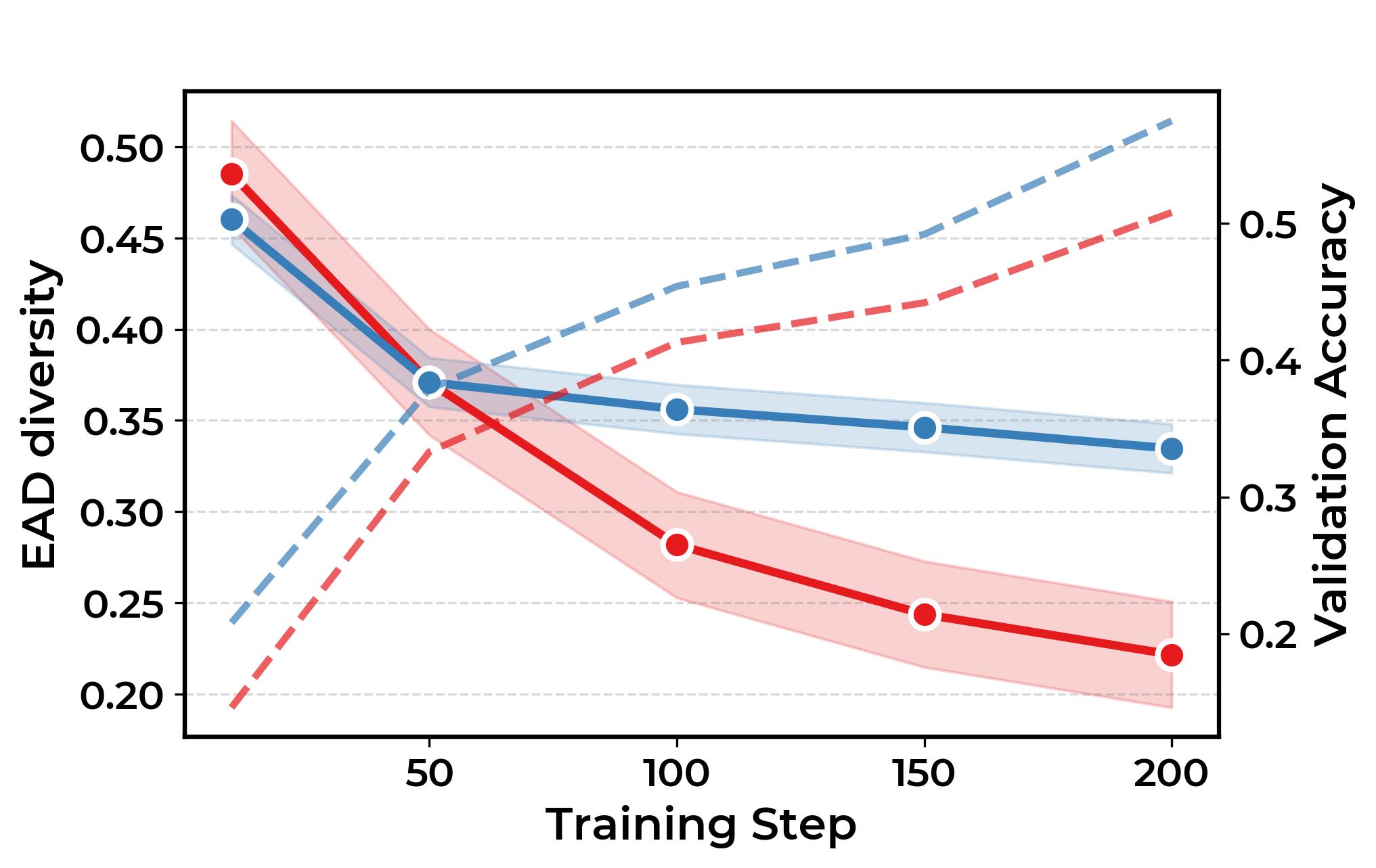}
        \caption{EAD}
        \label{fig:diversity_ead_per}
    \end{subfigure}
    \hfill
    \begin{subfigure}[t]{0.32\linewidth}
        \centering
        \includegraphics[width=\linewidth]{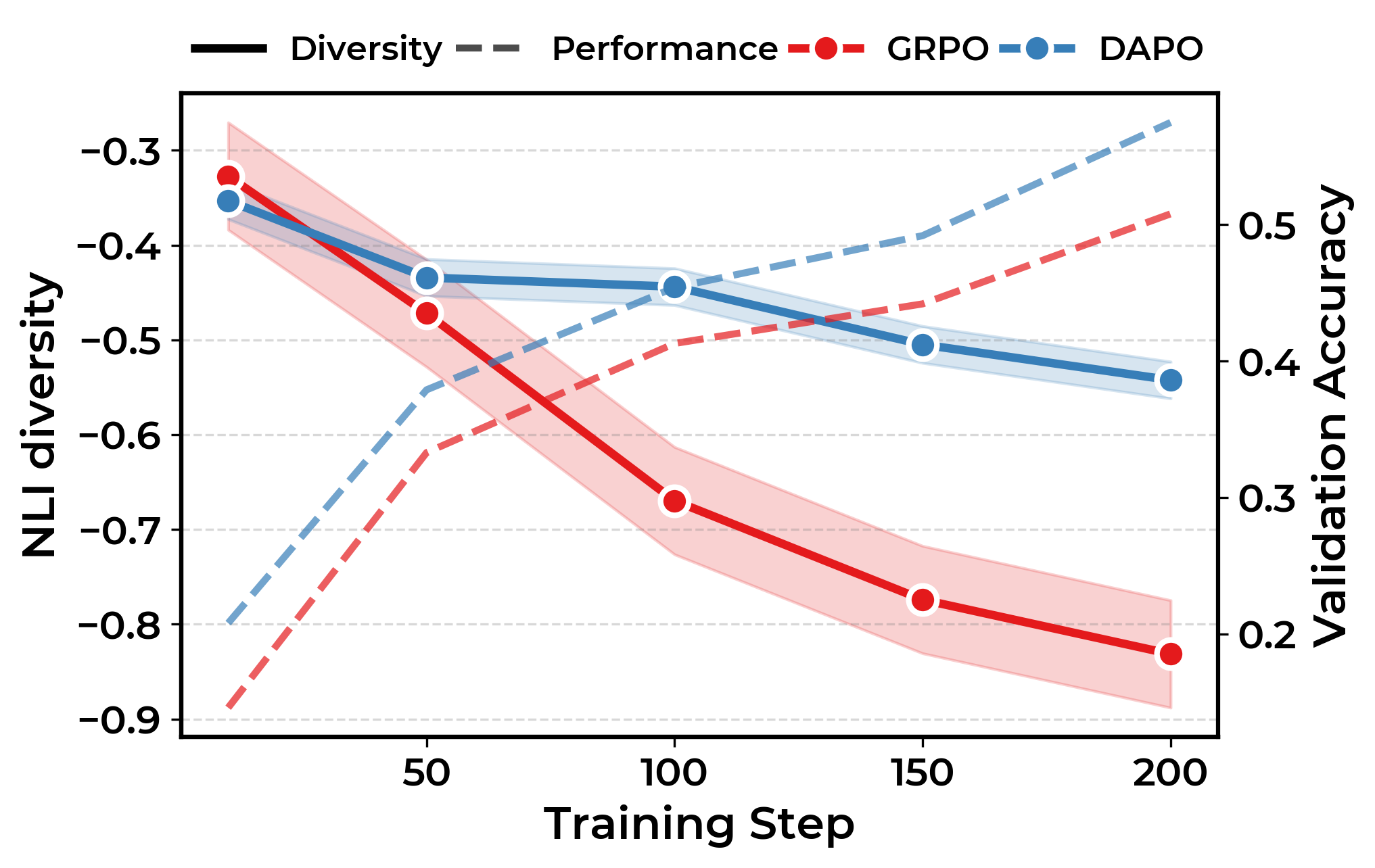}
        \caption{NLI}
        \label{fig:diversity_nli_per}
    \end{subfigure}
    \hfill
    \begin{subfigure}[t]{0.32\linewidth}
        \centering
        \includegraphics[width=\linewidth]{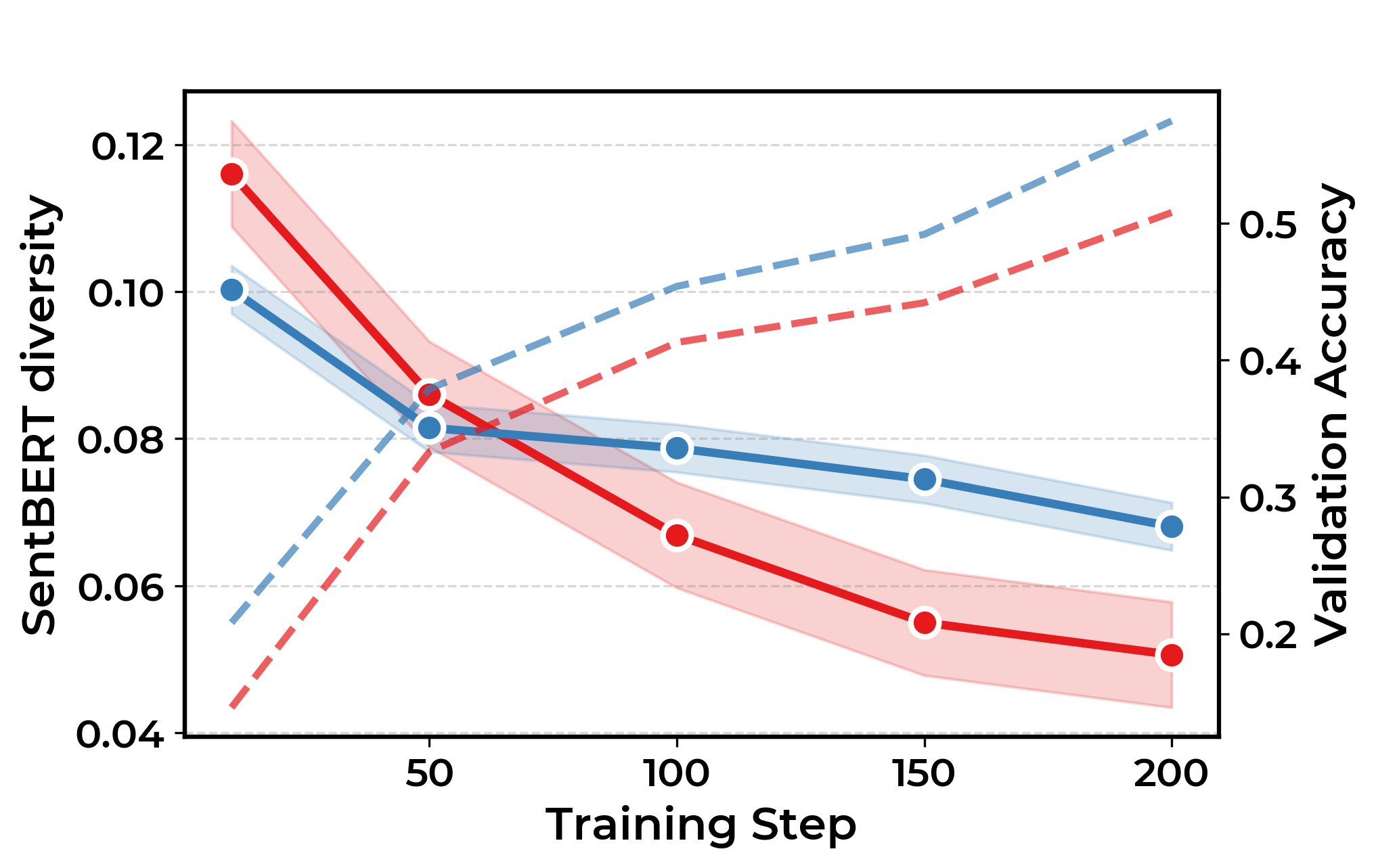}
        \caption{Embedding}
        \label{fig:diversity_embed_per}
    \end{subfigure}

    \caption{Evolution of generation diversity during RLVR training under DAPO and GRPO, measured in a per-input setting by three complementary metrics capturing lexical (EAD), logical (NLI), and semantic (embedding) diversity.}
    \label{fig:diversity_training_per}
\end{figure*}
\paragraph{Problem Definition: }
We study the problem of detecting whether a given example has appeared during the RLVR training stage of a reasoning language model. Formally, given an RLVR-tuned model $M_{\text{RLVR}}$ trained on an (unknown) RL dataset $D_{\text{RL}}$, the goal is to learn a detector $h$ that determines whether an arbitrary query input $x$ belongs to $D_{\text{RL}}$. The detector predicts membership as
\[
h(x, M_{\text{RLVR}}) \rightarrow \{0, 1\}.
\]
This formulation follows the standard membership inference attack (MIA) setting, but differs from pretraining and supervised fine-tuning scenarios due to the reward-driven and self-generative nature of RLVR.

\paragraph{RLVR Training Process: }
During RLVR training, each example consists of a reasoning-intensive prompt $x$ paired with a verifiable answer $a$. The model first generates a CoT $c$ and a final answer $\hat{a}$ sampled from its policy $\pi_\theta$, that is
$(c, \hat{a}) \sim \pi_\theta(\cdot \mid x)$.
The output of the model, denoted as $o$, is the combination of the CoT and the final answer: 
\[
o = (c, \hat{a}).
\]
A scalar reward $r(\hat{a}, a)$ is computed based on the correctness of the predicted answer. The model parameters are updated to increase the expected reward
$J(\theta) = \mathbb{E}_{(c,\hat{a}) \sim \pi_\theta(\cdot \mid x)}[\, r(\hat{a},a) \,]$,
and a typical policy gradient update takes the form
$\nabla_\theta J(\theta)
= \mathbb{E}_{(c,\hat{a}) \sim \pi_\theta(\cdot \mid x)}
\big[\, r(\hat{a},a)\,\nabla_\theta \log \pi_\theta(c,\hat{a}\mid x) \big].$

This optimization acts on the distribution of generated chains of thought rather than on the prompt itself, and no golden chain of thought is available for likelihood comparison. As a result, RLVR repeatedly adjusts the probability of reasoning trajectories according to their observed rewards. Unlike likelihood-based pretraining, RLVR optimizes models through reward feedback on self-generated reasoning trajectories, rather than optimizing on golden trajectories, making conventional token-level signals ineffective. This motivates a closer examination of how RLVR affects reasoning diversity.
\section{Analyzing Reasoning Pattern under RLVR}
Previous work has demonstrated that RLVR models tend to reduce the coverage of reasoning, leading to narrower reasoning trajectories compared to their base models~\cite{gandhi2025cognitive, yue2025does}. While these studies highlight the loss of reasoning diversity, they primarily focus on performance metrics rather than directly analyzing the changes in output structure. Notably, there has no systematic study quantifying how reasoning diversity evolves through the RLVR training process, especially for seen and unseen samples. To fill this gap, we focus on analyzing the dynamics of reasoning diversity in the outputs of RLVR models. We investigate how the diversity of reasoning trajectories decreases during RLVR training and explore how these changes can serve as a potential signal for detecting whether a specific example has been included in the training data. Specifically, we train a Qwen-2.5-7B-Base model~\cite{qwen2024qwen2} using two representative RLVR algorithms, DAPO~\cite{yu2025dapo} and GRPO~\cite{shao2024deepseekmath} (Details in App.~\ref{app:analysis_details}). The DAPO algorithm employs a clip-higher strategy to explicitly encourage exploration during training. We conduct RLVR training on the DAPO dataset~\cite{yu2025dapo} for 200 rollout steps and analyze multiple intermediate checkpoints. 

\paragraph{RLVR Induces Generation Rigidity.}
To quantify the impact of RLVR on output diversity, we use three complementary metrics, as proposed by \citet{kirk2023understanding}: lexical, logical, and semantic diversity. Lexical diversity is measured using expectation-adjusted distinct n-grams (EAD)~\cite{liu2022rethinking}, which estimates the proportion of distinct n-grams while correcting for length bias, averaged over $n = \{1,2,3,4,5\}$. In particular, for a given prompt, we calculate the distinct n-grams from completions and normalize them by the expected number of n-grams.  Logical diversity is assessed using an NLI-based metric, which measures the proportion of entailment and contradiction relations across pairs of randomly selected completions. This allows us to quantify how much the model’s reasoning varies in terms of logical consistency. Finally, semantic diversity is evaluated through sentence-embedding diversity, defined as one minus the average pairwise cosine similarity between the embeddings of the generated outputs. Together, these metrics capture lexical, logical, and semantic diversity, respectively, providing a comprehensive view of how RLVR affects different aspects of generation behavior.

For each checkpoint, we sample 32 completions for 300 training prompts and compute the per-input diversity across all three metrics. As shown in Fig.~\ref{fig:diversity_ead_per}, we observe a steady decline in all three metrics throughout RLVR training under both DAPO and GRPO. This consistent decrease suggests that the space of reasoning trajectories conditioned on a fixed prompt narrows as the model undergoes RLVR, indicating that RLVR systematically leads to converged reasoning trajectories in lexical, logical, and semantic dimensions. We further analyze the diversity in a cross-input setting, where we assess the diversity across completions from all 300 prompts (Fig.~\ref{fig:diversity_training_cross} in App.~\ref{app:cross_input_diversity}). We observe a similar declining trend in lexical diversity, further supporting the notion that RLVR reduces output diversity not only at the level of individual prompts but also across the global output distribution. This global reduction in diversity reflects the collapse of the reasoning space, concentrating it into a smaller set of recurring patterns, leading to more rigid outputs as training progresses.

\paragraph{RLVR Converges Symbolic Reasoning Segments.}
\begin{figure}[t]
    \centering
    \includegraphics[width=\linewidth]{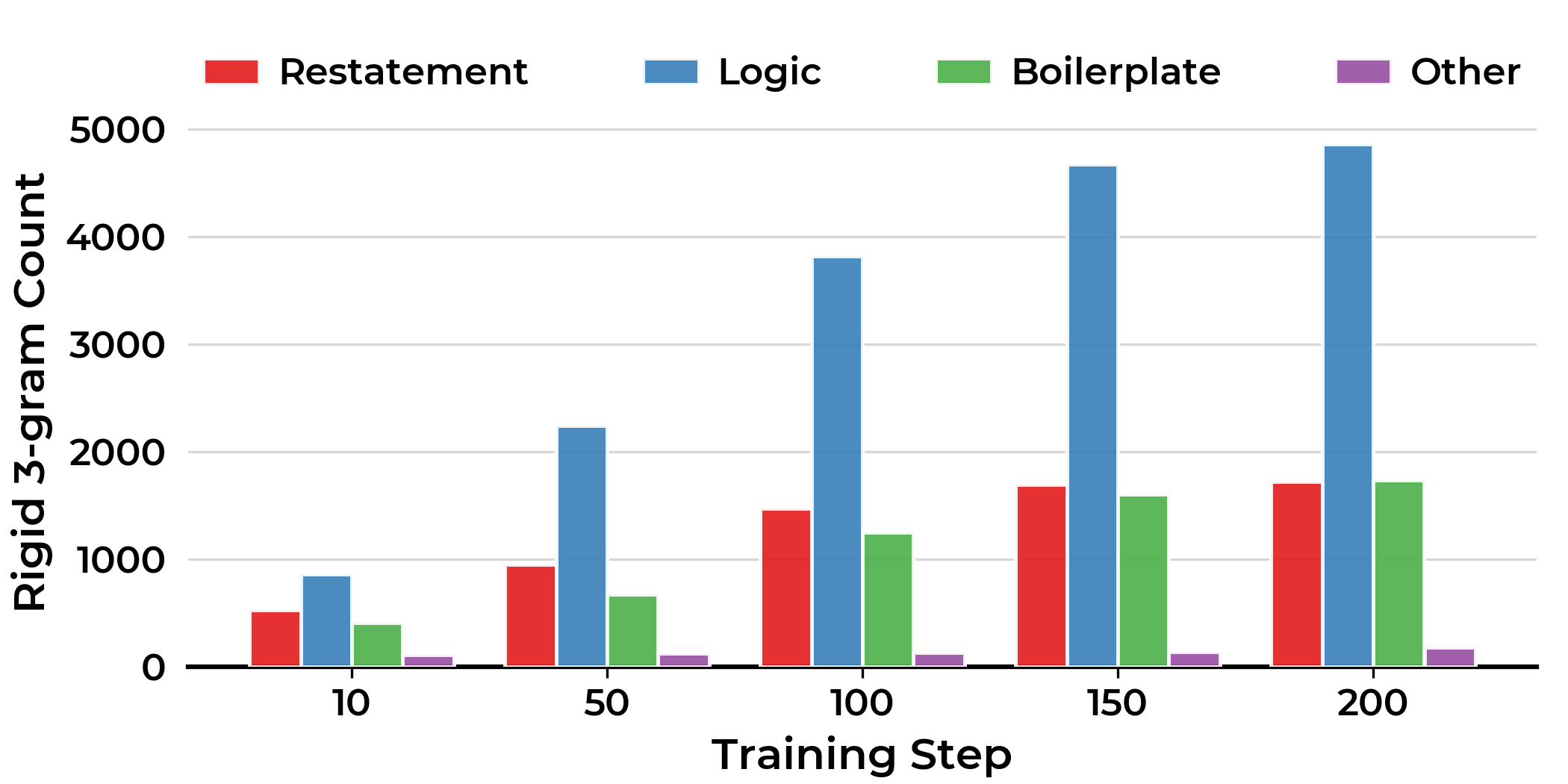}
    \caption{
    Rigid 3-gram category counts in GRPO training checkpoints. GPT-4o labeled rigid 3-grams into four categories.
    }
    \label{fig:category_counts}
\end{figure}
To understand which parts of the completions are most affected by RLVR training, we inspect 3-grams that appear in at least half of the completions for a given prompt. Through this analysis, we identify several repeated fragments can be categorized into three main types. The first type is restatements of the problem, where the model habitually repeats the problem statement in the early part of the output. The second type is boilerplate connective phrases, such as “To solve this problem...” or “Let's denote the...”. These phrases are structural fillers that do not contribute to the core reasoning but serve to connect the reasoning steps. The third type, symbolic or algebraic logic steps, represents the core reasoning components, typically involving algebraic manipulations, standardized mathematical transformations, or function definitions. For example, a model might express equations like “x = y + 2,” apply mathematical laws, or define functions like “f(x) = x\textasciicircum 2 + 3x,” all of which involve relatively fixed structural forms that allow little variation during reasoning. These segments play a critical role in the core of the model’s reasoning and tend to become more standardized as RLVR training progresses. For example, Fig.~\ref{fig:ngram_case} in App.~\ref{app:ngram_case} shows a representative case of these categories. 

We extend this observation by sampling 50 training prompts at each checkpoint and extracting the high-frequency 3-grams. These 3-grams are then categorized with GPT-4o(prompt is shown in App.~\ref{app:labeling_prompt}), allowing any unclassifiable 3-grams to be marked as “other.” The results are plotted for each checkpoint, showing the number of recurring 3-grams. As shown in Fig.~\ref{fig:category_counts}, we observe that symbolic logic fragments increase rapidly over the course of RLVR training, while restatement and boilerplate patterns exhibit slower growth. This indicates that RLVR significantly compresses the symbolic reasoning components, especially the logical tokens, which carry the model’s reasoning structure. As these segments become more rigid, they form a core set of standardized reasoning steps, leaving less variability in the model’s outputs. Thus, we conclude that RLVR primarily causes the collapse of symbolic reasoning segments.

\paragraph{RLVR Converges Reasoning into a Limited Set of Structural Modes.}
\begin{figure}[t]
    \centering
    \includegraphics[width=\linewidth]{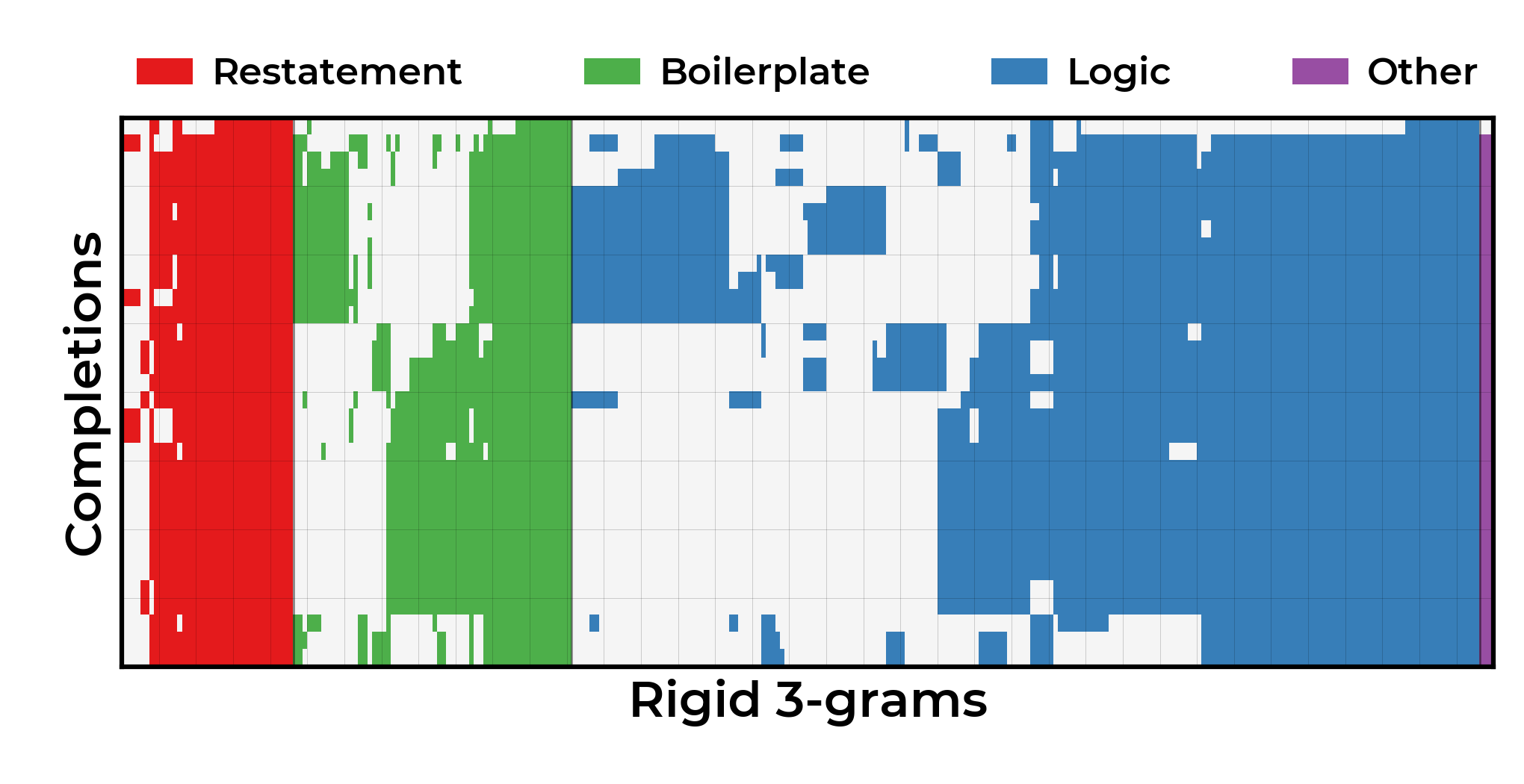}
    \caption{
    Heatmap of rigid 3-grams that appear in at least three completions for a single prompt. The clustering is achieved via hierarchical agglomerative clustering.
    }
    \label{fig:n_gram_heatmap}
\end{figure}
\begin{figure}[t]
    \centering
    \includegraphics[width=0.9\linewidth]{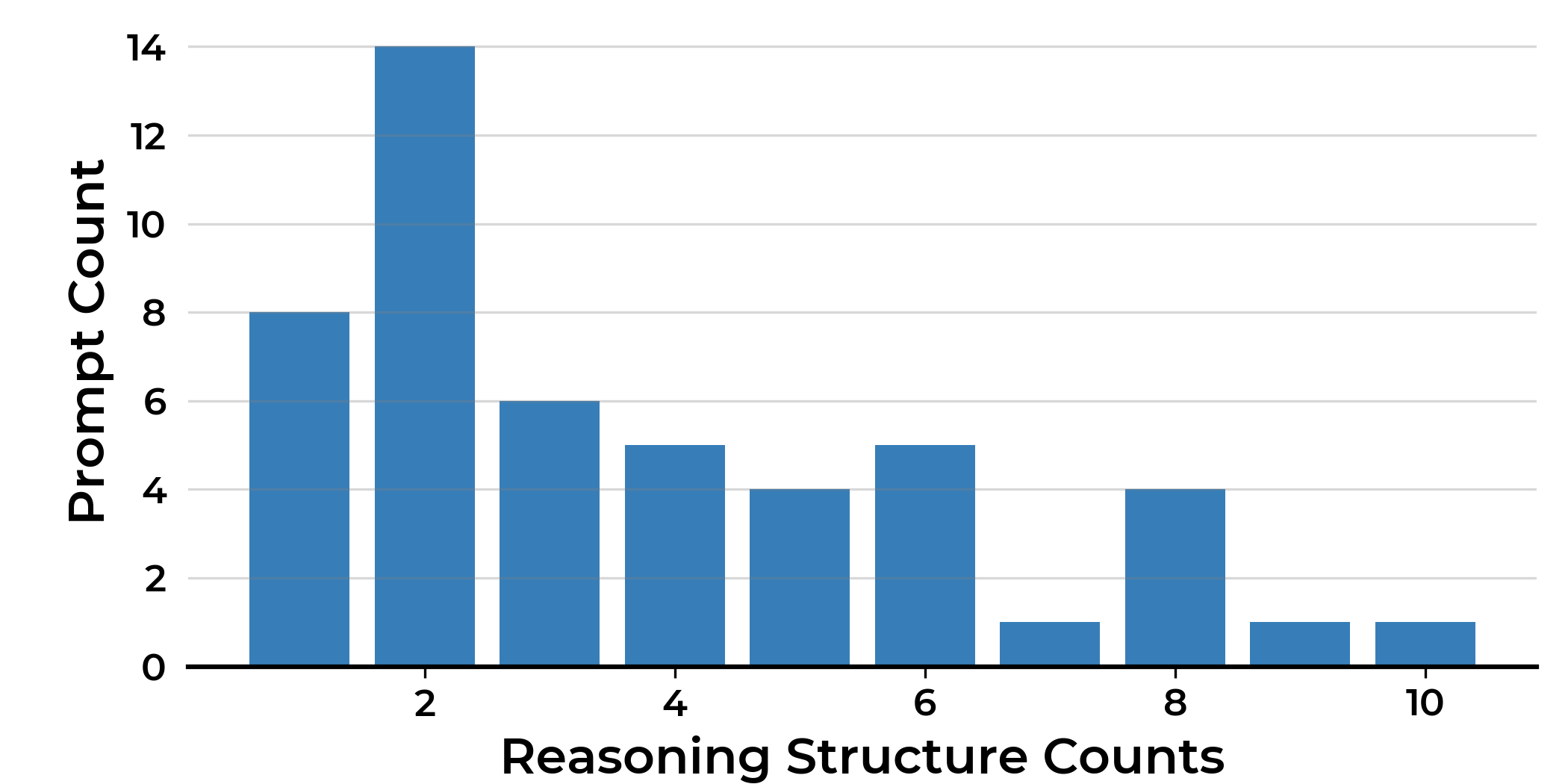}
    \caption{
    Distribution of the number of reasoning structure clusters across 50 training prompts.
    }
    \label{fig:logic_cluster_hist_train}
\end{figure}
To explore how RLVR affects reasoning trajectories, we analyze whether it leads to a single deterministic path or multiple recurring structural modes. As shown in Fig.~\ref{fig:n_gram_heatmap}, we observe that logical $3$-grams exhibit clustering patterns and often co-occur with boilerplate phrases. Restatements of the problem appear across all completions. Fig.~\ref{fig:overview} (middle) provides a qualitative example, where completions are grouped into distinct reasoning structures. Each structure reflects a similar solution path, showing the model’s tendency to collapse into a limited number of reasoning modes during RLVR training.

To quantify these patterns, we apply hierarchical agglomerative clustering~\cite{murtagh2012algorithms} to the logic $n$-grams extracted from 50 training prompts and analyze the diversity of the reasoning structures across different RLVR checkpoints. The histogram in Fig.~\ref{fig:logic_cluster_hist_train} shows that most prompts exhibit between two and four stable clusters of reasoning structures, with a few prompts showing even more. This result indicates that RLVR compresses reasoning into a limited number of tightly concentrated structural modes rather than a single canonical trajectory. The clustering analysis further supports the idea that RLVR does not lead to a single deterministic solution, but rather to a small set of rigid patterns, highlighting the convergence of reasoning trajectories.

\paragraph{Greater Rigidity in Seen Data Compared to Unseen Data.}
\begin{table}[t]
\footnotesize
\centering
\caption{
Comparison of reasoning structure collapse between seen and unseen data. The first part shows the distribution of rigid 3-gram categories, while the second part displays the cumulative distribution of the number of reasoning structure clusters across different stages of RLVR training.
}
\label{tab:train_test_comparison}
\setlength{\tabcolsep}{7pt}
\begin{tabular}{lcccc}
\toprule
\multicolumn{5}{c}{\textbf{Rigid Logic 3-gram Count}} \\
\midrule
\textbf{Step} & \textbf{50} & \textbf{100} & \textbf{150} & \textbf{200} \\
\midrule
Seen  & 856 & 2,240 & 3,817 & 4,673 \\
Unseen   & 840 & 1,684 & 2,639 & 3,476 \\
\midrule
\multicolumn{5}{c}{\textbf{Cumulative Distribution of Reasoning Structure Clusters}} \\
\midrule
\textbf{Cluster Size} & \textbf{$\leq 2$} & \textbf{$\leq 4$} & \textbf{$\leq 8$} & \textbf{$\leq 16$} \\
\midrule
Seen & 44.90\% & 67.35\% & 95.92\% & 100.00\% \\
Unseen  & 22.45\% & 55.10\% & 89.80\% & 100.00\% \\
\bottomrule
\end{tabular}
\end{table}
Finally, we compare the reasoning structure convergence between seen and unseen data. As shown in Tab.~\ref{tab:train_test_comparison}, both seen and unseen data exhibit reasoning convergence, but with notable differences. For the rigid 3-gram categories, the training set shows a greater increase in symbolic reasoning steps, indicating a more rigid model. In contrast, the unseen data exhibits a smaller increase in rigidity, suggesting a lower level of rigidity for unseen prompts. Regarding reasoning structure clusters, the cumulative distribution reveals that training prompts have a higher proportion of clusters with fewer reasoning structures, while unseen prompts retain more diverse reasoning paths. This suggests that RLVR induces stronger rigidity for seen data, while unseen prompts maintain more variability in reasoning.

\section{\ours : A Simple Black-Box Detector for RLVR Exposure}
Previous analyses have shown that RLVR training leads to a collapse in the structural components of CoT reasoning, with completions for seen prompts converging into a few tight clusters. This characteristic provides a basis for membership inference: completions for member prompts should exhibit a stronger tendency to cluster together than those for non-member prompts. Building on this observation, we introduce \ours, a simple black-box statistic that quantifies the degree of this clustering.

Given a prompt $x$, we sample $m$ completions from $M_{\text{RLVR}}$
\[
O(x) = \{o_1, o_2, \ldots, o_m\}, \qquad o_i \sim M_{\text{RLVR}}(\cdot \mid x),
\]
and compute the pairwise normalized edit distance between every pair:
\[
D_{ij} = d(o_i, o_j), \qquad 1 \le i,j \le m.
\]
where $d(\cdot,\cdot)$ denotes the normalized Levenshtein edit distance between two completions, 
defined as the minimum number of insertions, deletions, and substitutions required to transform 
one sequence into the other, normalized by the length of the longer sequence.
For each completion $o_i$, we define its nearest--neighbor distance:
\[
\mathrm{NN}_i(x)=\min_{j \ne i} D_{ij}.
\]

Let $\mathrm{NN}_{(1)}(x) \le \mathrm{NN}_{(2)}(x) \le \cdots \le \mathrm{NN}_{(m)}(x)$, be the sorted list of nearest-neighbor distances.  
We take the $k$ smallest values to form a set $\text{Min-}k(x)$ and define the detection score:
\[
\mathrm{Min\text{-}kNN}(x)
=
\frac{1}{k} \sum_{t=1}^{k} \mathrm{NN}_{(t)}(x).
\tag{1}
\label{eq:minknn}
\]

Because RLVR collapse forces completions of seen prompts into a few compact structural modes, we expect
\[
\mathrm{Min\text{-}kNN}(x_{\text{seen}})
<
\mathrm{Min\text{-}kNN}(x_{\text{unseen}}).
\]

A detector is obtained by thresholding $\mathrm{Min\text{-}kNN}(x)$. The method requires only black-box 
sampling access to the RLVR-tuned model and does not use the token log probabilities, or reference models. 
\section{Experiments}
\label{sec:exp}
\begin{table*}[t]
\centering
\caption{
AUC results for RLVR data detection across RL-tuned models.
\textbf{BB} indicates whether a method operates with sampling-only access. DS-KK and DS-SAT denote DeepSeek-Math-7B-Instruct models RLVR-trained under the K\&K and SAT settings, respectively;
QW-KK and QW-SAT denote QW-2.5-7B-Instruct models RLVR-trained under the same settings.
The best result is shown in \textbf{bold} and the second best is \underline{underlined}.
}
\label{tab:auc_results}
\small
\setlength{\tabcolsep}{3.5pt}
\renewcommand{\arraystretch}{1.15}
\begin{tabular}{l c cccccccc|c}
\toprule
& & \multicolumn{4}{c}{\textbf{Open-source RLVR Models}} 
& \multicolumn{4}{c}{\textbf{RL-MIA Models}} & \\
\cmidrule(lr){3-6} \cmidrule(lr){7-10}
\textbf{Method} & \textbf{BB} &
\textbf{DAPO} &
\textbf{JustRL} &
\textbf{ORZ} &
\textbf{SimpleRL} &
\textbf{DS-KK} &
\textbf{QW-KK} &
\textbf{DS-SAT} &
\textbf{QW-SAT} &
\textbf{Avg.} \\
\midrule
PPL~\cite{shi2023detecting}
& \xmark
& \underline{0.63} & \underline{0.61} & \underline{0.72} & \underline{0.72}
& \underline{0.59} & \underline{0.56} & 0.44 & \underline{0.54}
& \underline{0.60} \\
Min-K\%~\cite{shi2023detecting}
& \xmark
& 0.37 & 0.43 & 0.29 & 0.28
& 0.53 & 0.44 & 0.53 & 0.47
& 0.42 \\
Min-K\%++~\cite{zhang2024min}
& \xmark
& 0.37 & 0.48 & 0.56 & 0.39
& 0.56 & 0.53 & \underline{0.66} & 0.52
& 0.51 \\
CDD~\cite{dong2024generalization}
& \cmark
& 0.52 & 0.51 & 0.36 & 0.45
& 0.49 & 0.46 & 0.50 & 0.47
& 0.47 \\
Recall~\cite{xie2024recall}
& \xmark
& 0.50 & 0.52 & 0.51 & 0.62
& 0.54 & 0.52 & 0.56 & 0.47
& 0.53 \\
Self-Critique~\cite{tao2025detecting}
& \xmark
& 0.58 & 0.51 & 0.48 & 0.49
& 0.54 & 0.53 & 0.43 & 0.49
& 0.51 \\
\midrule
\rowcolor[HTML]{D9E7FA} 
\textbf{\ours}
& \cmark
& \textbf{0.72} & \textbf{0.66} & \textbf{0.75} & \textbf{0.76}
& \textbf{0.80} & \textbf{0.68} & \textbf{0.68} & \textbf{0.57}
& \textbf{0.70 (+17\%)} \\
\bottomrule
\end{tabular}
\vspace{-8pt}
\end{table*}
\subsection{Experimental Setup}
\paragraph{Models and Datasets}
To comprehensively evaluate the effectiveness of \ours\ , we consider two complementary evaluation settings: publicly released RLVR-trained models and controlled RL-MIA benchmarks.
For \textbf{Open-source RLVR models,} we evaluate on several publicly released reasoning models that are available only after RLVR training, including SimpleRL-32B~\cite{zeng2025simplerl}, DAPO-Qwen-32B~\cite{yu2025dapo}, JustRL-DeepSeek-1.5B~\cite{he2025justrl}, and Open-Reasoner-Zero-7B~\cite{hu2025open}. For each model, we sample 300 examples from its RL training data as member (seen) instances. To create non-member (unseen) instances, we collect 300 reasoning problems that do not appear in the RL data of any evaluated model. These problems are sourced from AIME 2024 (30 problems), AIME 2025 (30 problems), Beyond-AIME~\cite{bytedance_seed_2025_beyondaime} (100 problems), Omni-Math~\cite{gao2024omnimathuniversalolympiadlevel} (100 problems), and MATH-500~\cite{wang2024math} (40 problems). This balanced set provides a clean and fair benchmark for detecting RLVR data in real-world open-source models.
\textbf{RL-MIA benchmark models.}
Following the RL-MIA setup of~\cite{tao2025detecting}, we include controlled contamination settings. Specifically, we train Qwen2.5-7B-Instruct and DeepSeek-Math-7B-Instruct~\cite{shao2024deepseekmath} using the SAT and K\&K settings. In these settings, the provided training portions are treated as member data, and additional synthetic samples are constructed as non-member instances, following the original benchmark protocol. This setting allows us to evaluate detection performance under precisely controlled RL contamination scenarios. See App.~\ref{app:exp_details} for details.

\paragraph{Baselines}
We compare \ours\ against several membership inference methods, most of which were originally proposed for pretraining data detection: (1) PPL, which uses perplexity as a proxy for memorization; (2) Min-K\% Prob~\cite{shi2023detecting}, which detects low-probability outlier tokens; (3) Min-K\%++~\cite{zhang2024min}, an improved extension of Min-K\%; (4) Recall~\cite{xie2024recall}, which measures likelihood shifts under unrelated prefixing; and (5) CDD~\cite{dong2024generalization}, which compares the edit distance between stochastic generations and greedy sampling, checking for sharp peaks in the distance distribution. Finally, (6) Self-Critique~\cite{tao2025detecting} is specifically designed for RL data contamination detection. It compares token-level entropy sequences between a model’s initial response and its self-critique, where high similarity in entropy is interpreted as evidence of contamination.

\paragraph{Implementation and Evaluation Metrics}
For each prompt, \ours\ samples 32 completions from the RLVR-tuned model, computes all pairwise edit distances among the generated outputs, and averages the $k$ smallest distances to obtain the final score. Unless otherwise stated, we set $k=10$, which we find to be stable across models and training regimes. The baseline experimental setup follows the methodology in~\citet{tao2025detecting}. Other details are shown in App.~\ref{app:implementation_details}.

We evaluate detection performance using the Area Under the ROC Curve (AUC), a threshold-independent metric widely adopted in membership inference and data contamination studies~\cite{shi2023detecting, tao2025detecting, zhang2024min}. AUC measures the probability that the detector assigns a higher score to a member example than to a non-member example, with 50\% corresponding to random guessing. 

\subsection{Main Results}
Tab.~\ref{tab:auc_results} reports AUC performance across all models. We also conduct cost analysis in App.~\ref{app:cost_analysis}.

\textbf{\ours\ consistently outperforms all baseline methods.}
\ours\ achieves the highest AUC across all evaluated models, with an average score of 0.70, which represents a 17\% relative improvement over the strongest baseline. In contrast, probability-based and consistency-based baselines often exhibit unstable or near-random performance.

\textbf{\ours\ is robust across different RLVR algorithms.}
\ours\ consistently performs well across models trained with different RLVR algorithms, including GRPO (SimpleRL-32B), DAPO (DAPO-Qwen-32B), and PPO (Open-Reasoner-Zero-7B)~\cite{schulman2017proximal}. This stability suggests that the signal detected by \ours\ is not dependent on the RL algorithm.

\textbf{\ours\ generalizes across model scales.}
From 1.5B to 32B parameters, \ours\ consistently maintains strong detection performance, despite significant variations in model size and training dynamics. Notably, \ours\ operates in a fully black-box setting, consistently outperforming methods that require access to token probabilities or internal likelihoods. These results demonstrate that RLVR-induced structural collapse provides a strong, model-agnostic signal for exposure detection.

\subsection{Ablation}
\begin{figure}[t]
    \centering
    \begin{minipage}[t]{0.32\linewidth}
        \centering
        \includegraphics[width=\linewidth]{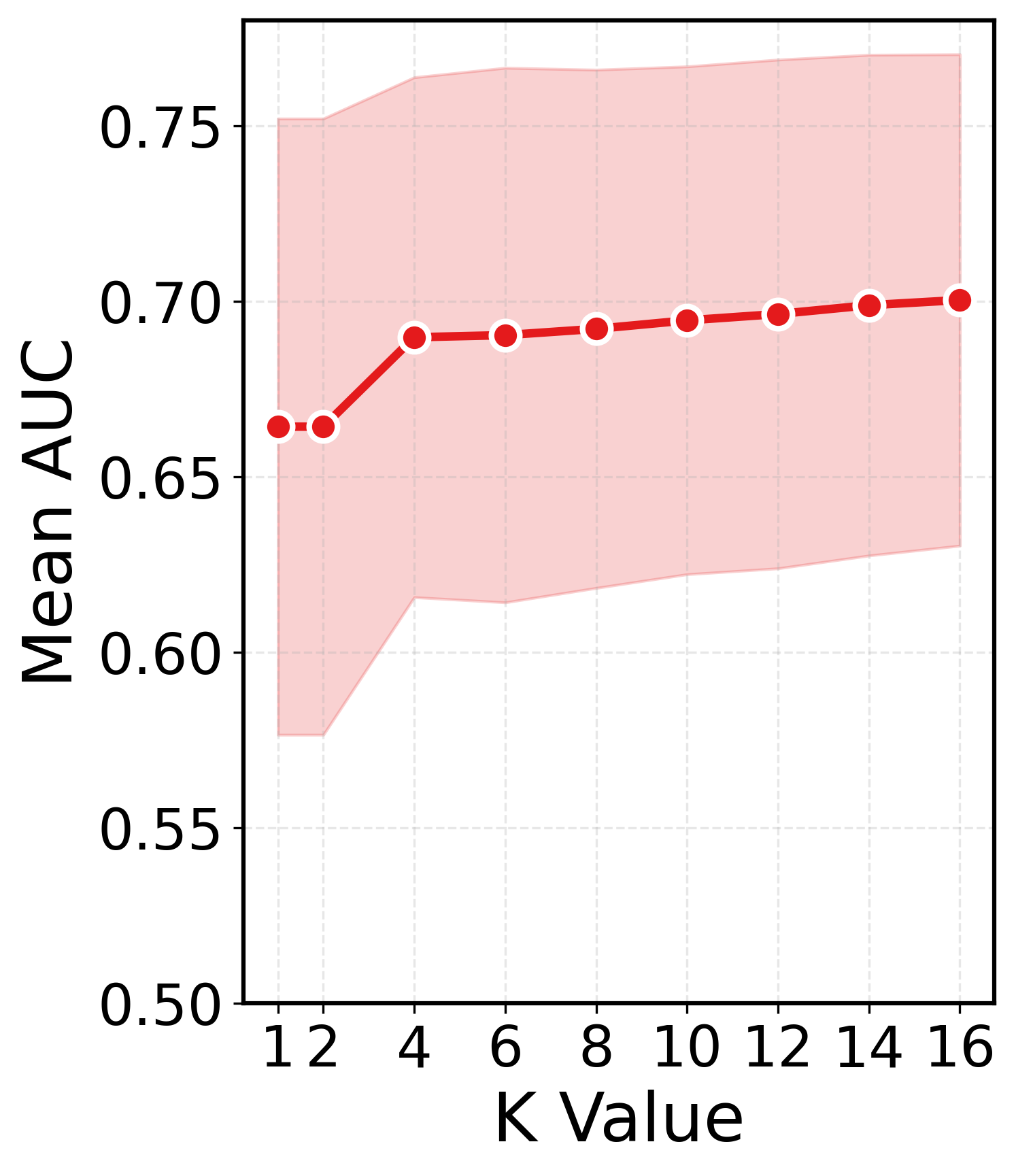}
    \end{minipage}
    \hfill
    \begin{minipage}[t]{0.32\linewidth}
        \centering
        \includegraphics[width=\linewidth]{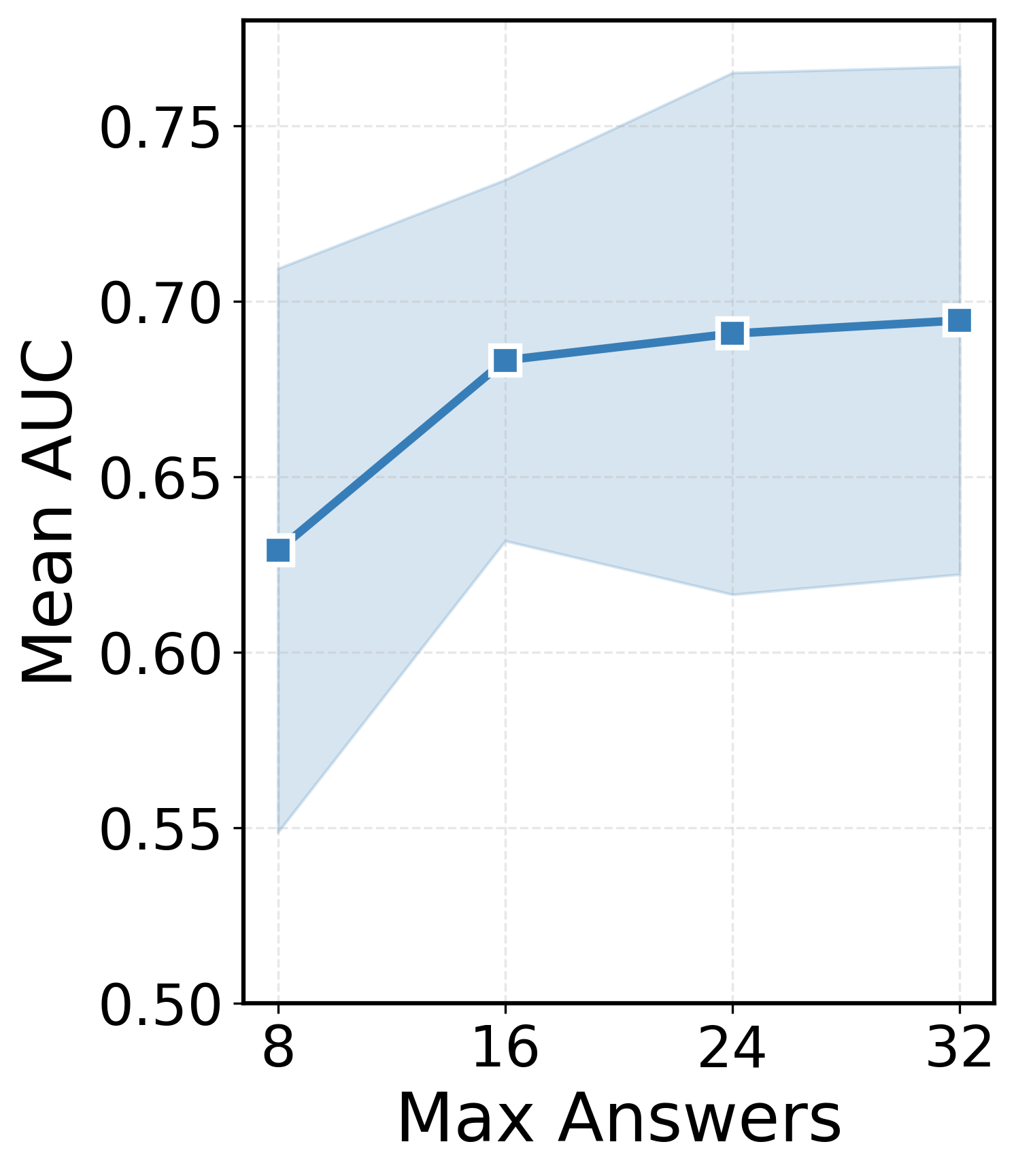}
    \end{minipage}
    \hfill
    \begin{minipage}[t]{0.32\linewidth}
        \centering
        \includegraphics[width=\linewidth]{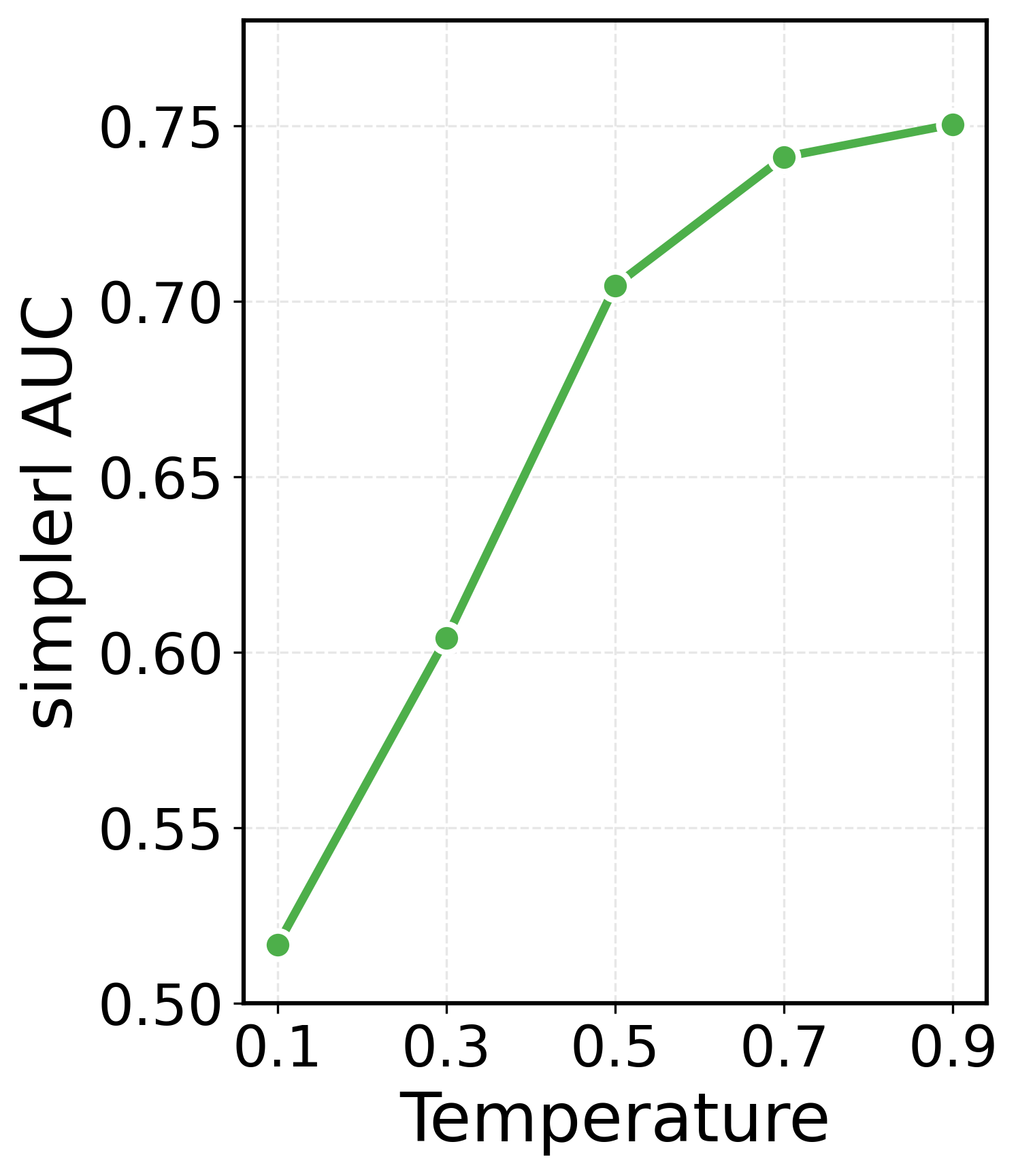}
    \end{minipage}

    \vspace{0.5em}
    \caption{
    Ablation studies of \ours .
    (a) Effect of $k$.
    (b) Effect of the number of sampled completions.
    (c) Effect of sampling temperature.
    }
    \label{fig:ablation}
\end{figure}
We analyze the sensitivity of \ours\ to its main hyperparameters.
For $k$ and the number of sampled completions, we report mean AUC over eight RLVR-trained models, while temperature is varied on SimpleRL setting.

\textbf{\ours\ is robust to the choice of $k$ over a wide range.}
As shown in the left panel of Fig.~\ref{fig:ablation}, performance improves quickly as $k$ increases from very small values and then saturates, with noticeable degradation only when $k$ is too small to reliably capture collapsed modes. 

\textbf{Sampling more completions consistently improves detection performance.}
The middle panel of Fig.~\ref{fig:ablation} demonstrates steady AUC gains as the number of sampled completions increases. This is because additional samples make collapsed structural patterns easier to expose, though the improvement plateaus after a certain point.

\textbf{Higher decoding temperatures strengthen the detection signal.} As illustrated in the right panel of Fig.~\ref{fig:ablation}, higher temperatures yield stronger detection performance, while lower temperatures reduce output variability and partially obscure structural collapse. 
We also observe that models trained with more extensive RLVR tend to require smaller $k$ to reach stable performance, whereas lower temperatures make \ours\ more sensitive to $k$.

\subsection{Analysis}
\paragraph{Robustness to Paraphrased Prompts.}
\begin{table}[t]
\centering
\caption{
Detection performance before and after paraphrasing RLVR training prompts under the DAPO setting.
}
\label{tab:paraphrase_analysis}
\footnotesize
\setlength{\tabcolsep}{6pt}
\renewcommand{\arraystretch}{1.15}
\begin{tabular}{lcc}
\toprule
\textbf{Method} & \textbf{Original} & \textbf{Paraphrased} \\
\midrule
PPL           & 0.63 & 0.66 \\
Self-Critique & 0.58 & 0.57 \\
\ours         & \textbf{0.72} & \textbf{0.71} \\
\bottomrule
\end{tabular}
\end{table}
In practice, a detector should remain effective when training prompts are paraphrased rather than exactly repeated. We therefore evaluate the robustness of \ours\ under the DAPO-Qwen-32B setting by paraphrasing 300 RLVR training prompts using GPT-4o and testing detection performance on these paraphrased inputs.
As shown in Tab.~\ref{tab:paraphrase_analysis}, \ours\ remains highly stable under paraphrasing, with AUC decreasing only slightly from 0.72 to 0.71, and other baselines exhibit similar changes. This indicates that the signal exploited by \ours\ is not tied to surface-level prompt forms, but instead reflects a structural collapse in reasoning induced by RLVR that generalizes across semantically equivalent prompts.

\paragraph{Detecting Distillation Prompts.}
\begin{table}[t]
\centering
\footnotesize
\setlength{\tabcolsep}{6pt}
\renewcommand{\arraystretch}{1.1}
\caption{Detection performance on distillation prompts (OpenR1-Distill-7B).}
\label{tab:distill_analysis}
\begin{tabular}{l c}
\toprule
\textbf{Method} & \textbf{AUC} \\
\midrule
PPL & 0.70 \\
Self-Critique & 0.65 \\
\ours & \textbf{0.76} \\
\bottomrule
\end{tabular}
\end{table}
Recent reasoning models are often distilled from RLVR-tuned models, where the RLVR-trained teacher model generates multiple reasoning trajectories, and the student is trained on these outputs. We analyze whether prompts used during distillation can be detected.
We conduct this analysis on OpenR1-Distill-7B~\cite{openr1}. We evaluate detection performance on two sets of prompts: 300 held-out evaluation problems and 300 prompts used during the distillation process. Using the same detection setup as in our main experiments, \ours\ achieves an AUC of 0.76 on distillation prompts, outperforming other baselines (Tab.~\ref{tab:distill_analysis}). These results suggest that the structural collapse induced by RLVR is partially transferred to the distilled model, allowing \ours\ to detect prompts used during distillation.

\paragraph{Robustness across Different RL Data Sources.}
\begin{table}[t]
\centering
\caption{
Detection performance on code and math data under ProRL setting.
}
\label{tab:data_source_analysis}
\footnotesize
\setlength{\tabcolsep}{6pt}
\renewcommand{\arraystretch}{1.15}
\begin{tabular}{lcc}
\toprule
\textbf{Method} & \textbf{Math} & \textbf{Code} \\
\midrule
PPL           & 0.40 & 0.36 \\
Self-Critique & 0.75 & 0.47 \\
\ours         & \textbf{0.80} & \textbf{0.69} \\
\bottomrule
\end{tabular}
\end{table}
We further analyze the robustness of \ours\ across different types of RLVR training data. We conduct this analysis on Nemotron-Research-Reasoning-Qwen-1.5B~\cite{liu2025prorl}, which is trained with a mixture of math, coding data during RLVR.
From its RL training set, we sample 100 prompts from math and 100 from code as member examples. To construct unseen data, we use 100 problems from Beyond-AIME for math and 100 validation examples from Eurus Coding~\cite{cui2025process} for code.
As shown in Tab.~\ref{tab:data_source_analysis}, \ours\ achieves strong performance on both math (AUC 0.80) and coding (AUC 0.69), outperforming other methods. However, RL-phase contamination detection is more challenging for code, likely due to the higher diversity and flexibility in coding tasks compared to the more structured nature of math problems.

\paragraph{Dual-Stage Contamination Analysis}
\begin{figure}[t]
    \centering
    \includegraphics[width=\linewidth]{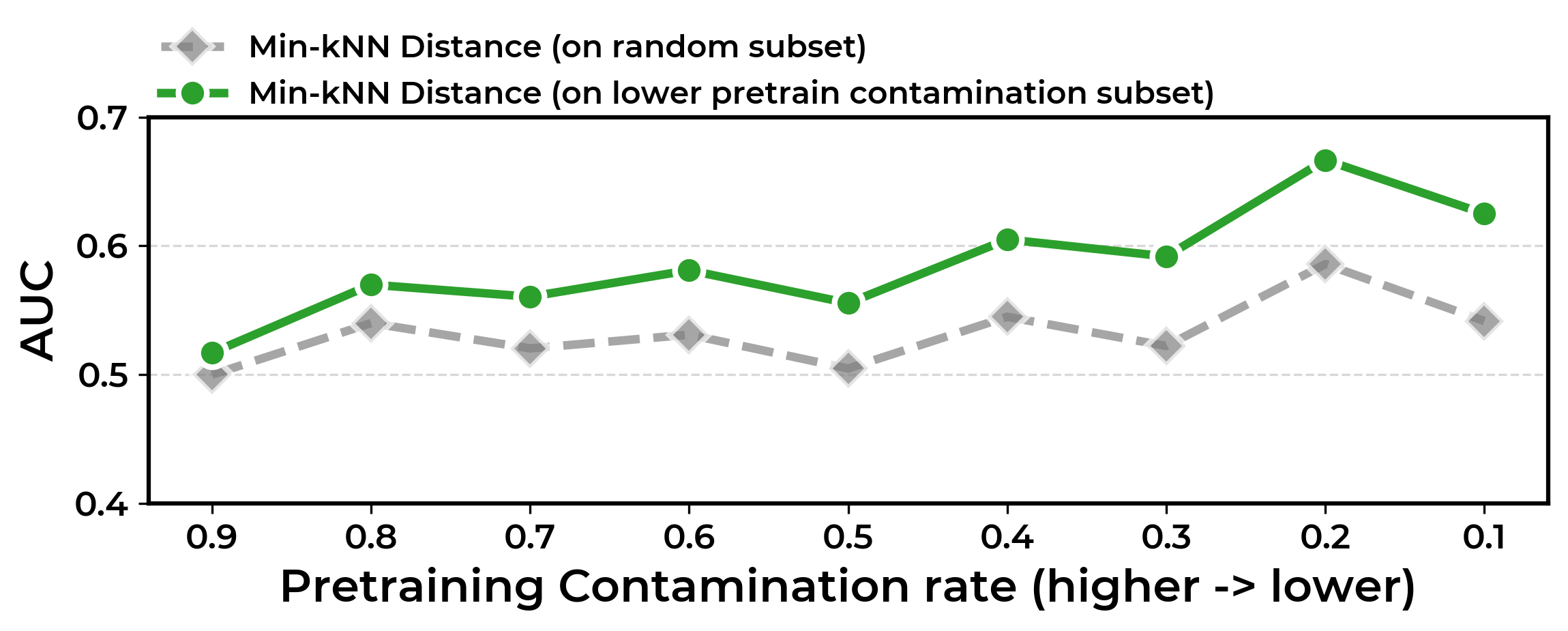}
    \caption{
        Dual-stage contamination analysis. The green line shows \ours\ RL data detection performance on the lower pretraining contamination subset, selected based on the bottom-q quantile of PPL scores. The gray line shows performance on the random-control subset, matched in size to control for sample size effects on AUC.
    }
    \label{fig:dual_contamination_analysis}
\end{figure}
We follow the dual-stage contamination detection experiment setup from \citet{tao2025detecting}, using the GSM8K dataset~\cite{cobbe2021training}, which is known for its pretraining contamination. RL-phase contamination is simulated by injecting a portion of the test set into the RL training process and fine-tuning with PPO. Two data subsets are created for contamination detection. The first subset, representing lower pretraining contamination, is created by selecting the bottom portion of items based on pretraining-contamination scores (using PPL). The second subset is a random-control subset, sampled uniformly to match the sample size of the first subset, controlling for sample size effects on AUC.
As shown in Fig.~\ref{fig:dual_contamination_analysis}, we compare RL-phase contamination effects across these subsets. The green line represents performance on the lower pretraining contamination subset, while the gray line shows performance on the random-control subset. The results indicate that \ours\ performs significantly better on the lower pretraining contamination subset, while the random-control subset shows reduced performance. This suggests that \ours\ is more effective at detecting RL-phase contamination when pretraining contamination is lower, validating our hypothesis that RL-phase contamination is easier to detect in data with weaker pretraining contamination signals. Moreover, \ours\ can be used alongside common pretraining detection methods to help determine the training phase in which the data appeared.
\section{Related Work}
\paragraph{Detect Data Exposure in Language Models.}
A growing body of work investigates whether large language models have been trained on specific data, driven by concerns about benchmark contamination, privacy leakage, and evaluation reliability. Most existing methods focus on pretraining or supervised fine-tuning, where data exposure shows up as statistical signals like low perplexity or probability outliers~\cite{shi2023detecting,dong2024generalization,zhang2024min,xie2024recall,carlini2021extracting,mireshghallah2022empirical,fu2023practical}. Reinforcement learning post-training introduces a different challenge. In RLVR, models are optimized via reward feedback on self-generated reasoning, making likelihood-based methods less effective. ~\citet{tao2025detecting} proposed self-critique to detect RL data contamination by identifying entropy collapse. Our work complements this by showing that RLVR induces a collapse in reasoning trajectory diversity for seen prompts, using structural similarity across multiple generations for RLVR data exposure detection, with a fully black-box approach.

\paragraph{Effects of Reinforcement Learning on Generation Diversity.}

Reinforcement learning with preference or alignment objectives consistently reduces generation diversity compared to supervised fine-tuning or non-preference methods. This effect has been observed across lexical, syntactic, semantic, and conceptual dimensions, indicating a compression of the model’s output space~\cite{kirk2023understanding,perez2022red,slocum2025diverse,murthy2025one}. Studies show that this reduction in diversity is linked to decreased creativity and exploratory behavior. For example, \citet{mohammadi2024creativity} show that RLHF lowers token-level entropy and embedding-space diversity, driving models toward fewer attractor states, which negatively impacts open-ended generation tasks. \citet{casper2023open} discuss mode collapse in RL-trained models, attributing it to reward maximization. More recent work, like \citet{murthy2025one}, shows that RLHF and RLAIF methods reduce a model’s ability to capture conceptual diversity in human responses, even with high surface-level quality. These findings suggest that diversity reduction is not merely a trade-off but a fundamental shift caused by reinforcement-based alignment. Unlike prior work, which focuses on diversity collapse as a trade-off in alignment, our study examines how this collapse manifests structurally in reasoning trajectories and uses it to detect RLVR training data exposure.
\section{Conclusion}
We investigate training data exposure in RL-post-trained reasoning models, identifying a structural collapse in reasoning trajectories during RLVR and proposing \ours, a black-box detection method leveraging this. \ours\ requires no access to training data or model internals. Experiments across various models and RL algorithms show \ours consistently outperforms baselines, providing a robust signal for contamination detection. Limitations of our work are outlined in App.~\ref{app:limitations}.

\section*{Impact Statement}
This work aims to improve the detection of RLVR exposure in AI models, a crucial step toward increasing transparency in machine learning systems. By enabling better detection of data contamination, our method can help ensure that models are more reliable and fair in their evaluations. As AI continues to play a larger role in critical applications, understanding and mitigating potential biases is essential. While our focus is on detection techniques, we recognize the importance of ethical considerations such as privacy and fairness, and we believe our work contributes to these broader conversations. At this time, we do not identify any specific ethical risks that require immediate attention.


\bibliography{main}
\bibliographystyle{icml2026}

\newpage
\appendix
\onecolumn
\section{Analysis Details}
\subsection{Training Details}
\label{app:analysis_details}
For the analysis model, we train Qwen-2.5-7B-Base using the GRPO and DAPO algorithms on the DAPO training data. The training hyperparameters are shown in Tab.~\ref{tab:analysis_training_params}. The model is trained for a total of 200 rollout steps, which is approximately equivalent to 6 epochs.
\begin{table}[h]
\centering
\caption{
Hyperparameters used for training Qwen-2.5-7B-Base on GRPO and DAPO.
}
\label{tab:analysis_training_params}
\begin{tabular}{lc}
\toprule
\textbf{HyperParameter} & \textbf{GRPO/DAPO}\\
\midrule
Actor learning rate & $1.0 \times 10^{-6}$ \\
Train batch size & 512 \\
Max prompt length & 2048\\
Max generation length & 20480\\
Temperature & 1.0  \\
Samples per prompt & 16 \\
Tensor model parallel & 4 \\
Mini batch size  & 32 \\
Use KL loss & No \\
\bottomrule
\end{tabular}
\end{table}
\subsection{Cross-Input Generation Diversity}
\label{app:cross_input_diversity}
\begin{figure*}[h]
    \centering
    \begin{subfigure}[t]{0.32\linewidth}
        \centering
        \includegraphics[width=\linewidth]{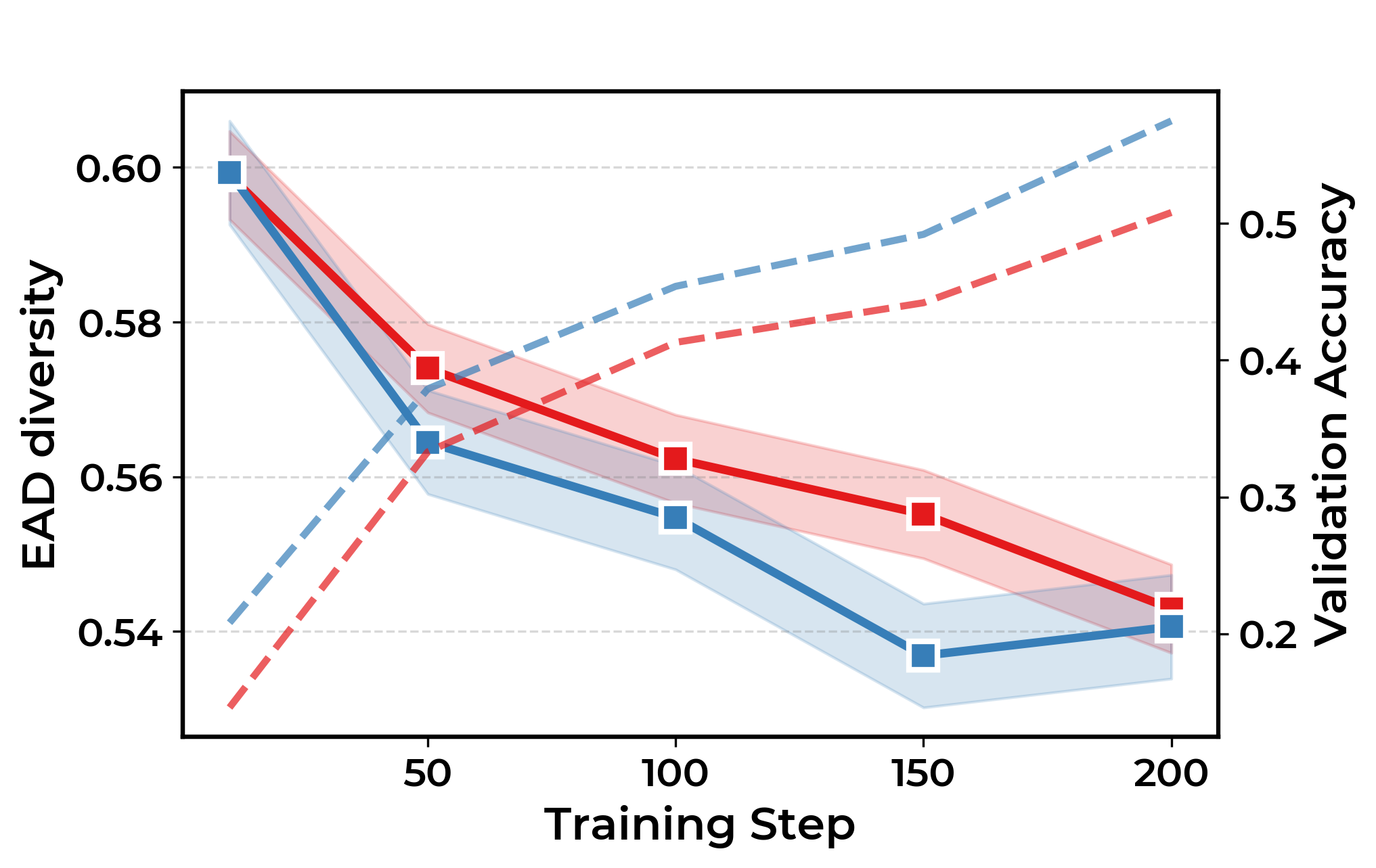}
        \caption{EAD}
        \label{fig:diversity_ead_cross}
    \end{subfigure}
    \hfill
    \begin{subfigure}[t]{0.32\linewidth}
        \centering
        \includegraphics[width=\linewidth]{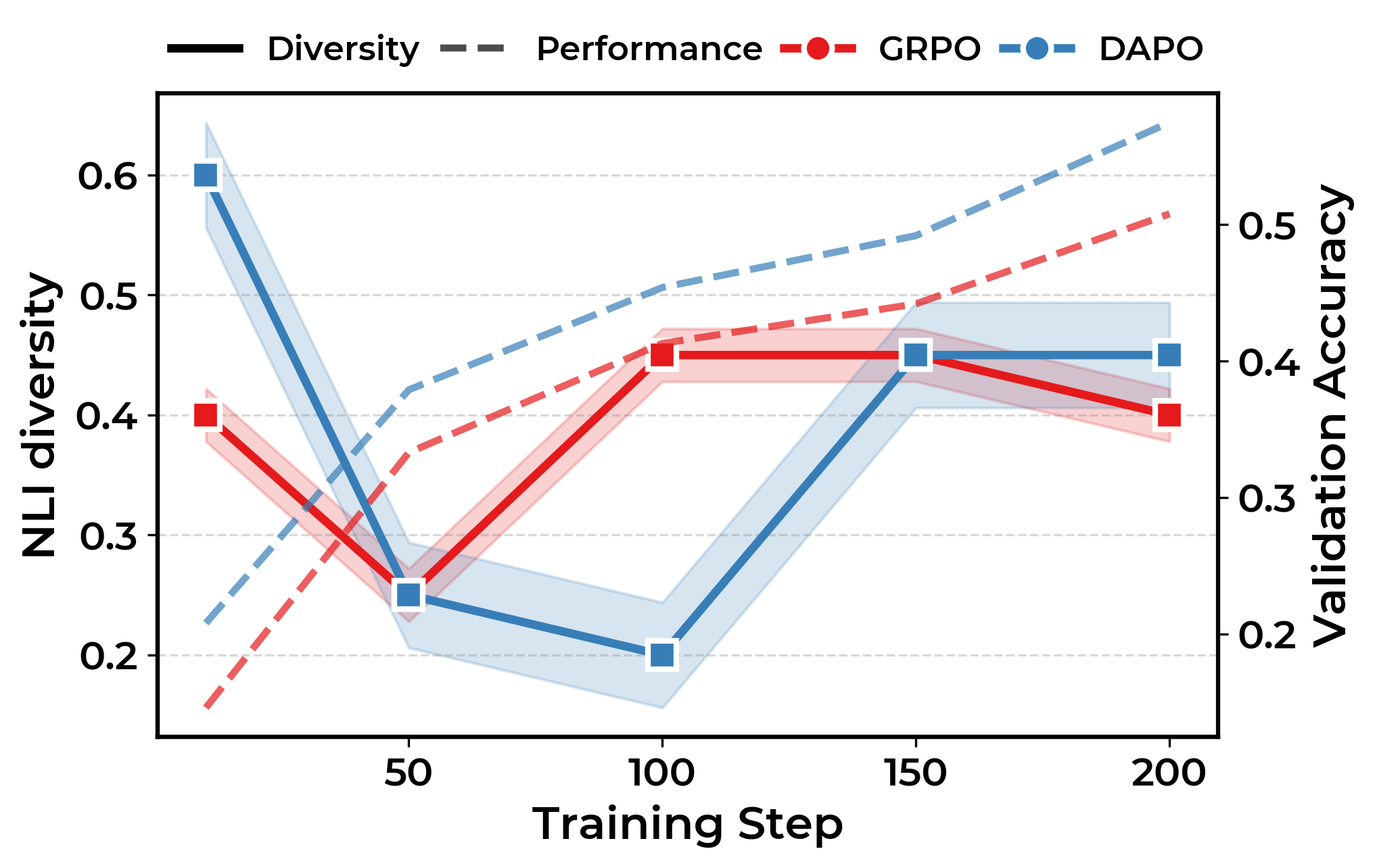}
        \caption{NLI}
        \label{fig:diversity_nli_cross}
    \end{subfigure}
    \hfill
    \begin{subfigure}[t]{0.32\linewidth}
        \centering
        \includegraphics[width=\linewidth]{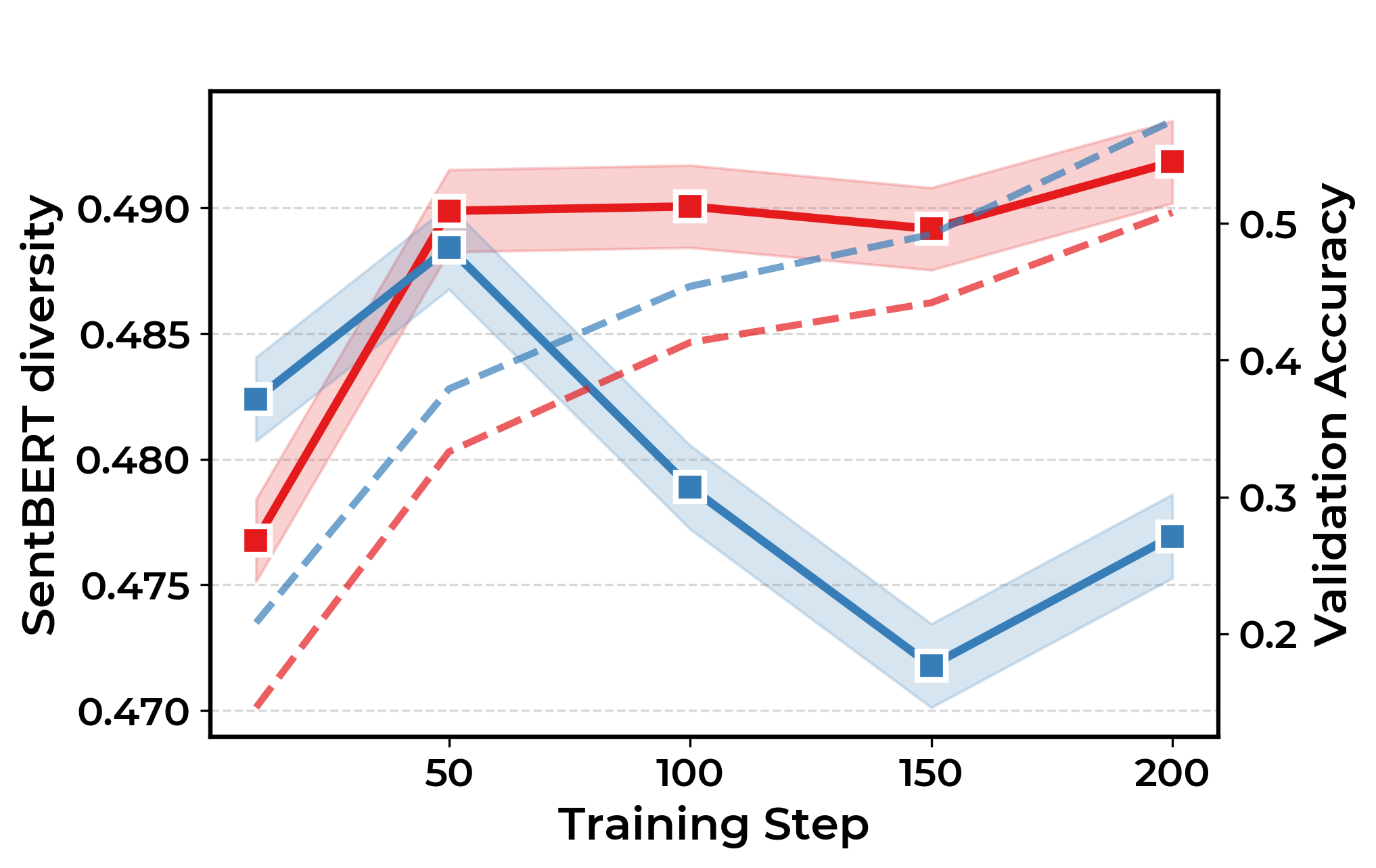}
        \caption{Embedding}
        \label{fig:diversity_embed_cross}
    \end{subfigure}

    \caption{Evolution of generation diversity during RLVR training under DAPO and GRPO, measured in a cross-input setting by three complementary metrics capturing lexical (EAD), logical (NLI), and semantic (embedding) diversity.}
    \label{fig:diversity_training_cross}
\end{figure*}
\newpage
\subsection{Prompt For Labeling}
\label{app:labeling_prompt}
\begin{tcblisting}{
  listing only,
  listing engine=listings,
  enhanced, breakable,
  colback=blue!6, colframe=blue!60!black,
  arc=3mm, boxrule=0.6pt,
  left=6mm, right=6mm, top=4mm, bottom=4mm,
  title={Labeling Rigid N-gram Prompt},
  % --- 下面是建议增加的关键配置 ---
  listing options={
    language=Python,
    basicstyle=\ttfamily\small,
    breaklines=true,         % 必须：否则长字符串会超出框外
    breakatwhitespace=true,  % 必须：确保在空格处折行，不破坏单词
    columns=fullflexible,
    showstringspaces=false   % 隐藏代码中字符串的空格符号
  }
}
{
    "task": (
        "You will be given: (1) a problem statement, (2) one sample model answer for that problem, and (3) a list of n-grams extracted from multiple model answers to the same problem. Your job is to label EACH n-gram with exactly one category from: restatement, logic, boilerplate, other.
        
        Labeling rules:
        1) restatement: the n-gram repeats or paraphrases the problem statement. If the n-gram appears verbatim in the problem statement, it MUST be labeled restatement.
        2) logic: the n-gram expresses problem-specific reasoning, math relations, formulas, constraints, or derived quantities.
        3) boilerplate: generic reasoning template language (e.g., 'we need to', 'let us', 'therefore', 'in conclusion'), or domain-agnostic filler.
        4) other: everything else that does not fit the above.
        
        Important:
        - Use the problem statement and the sample answer as context.
        - The sample answer includes inline markers [[...]] around matching n-grams.
        - Do not invent new n-grams or change the provided strings.
        - Output ONLY a JSON object mapping each input n-gram to one of the labels.
        - Include all n-grams, even if uncertain.
        - Keep the n-gram strings exactly as given."
    ),
    "problem": problem,
    "sample_answer": annotated_answer,
    "ngrams": ngrams,
}
\end{tcblisting}

\newpage
\subsection{Case on Rigidity}
\label{app:ngram_case}
\begin{figure}[H]
    \centering
    \includegraphics[width=\linewidth]{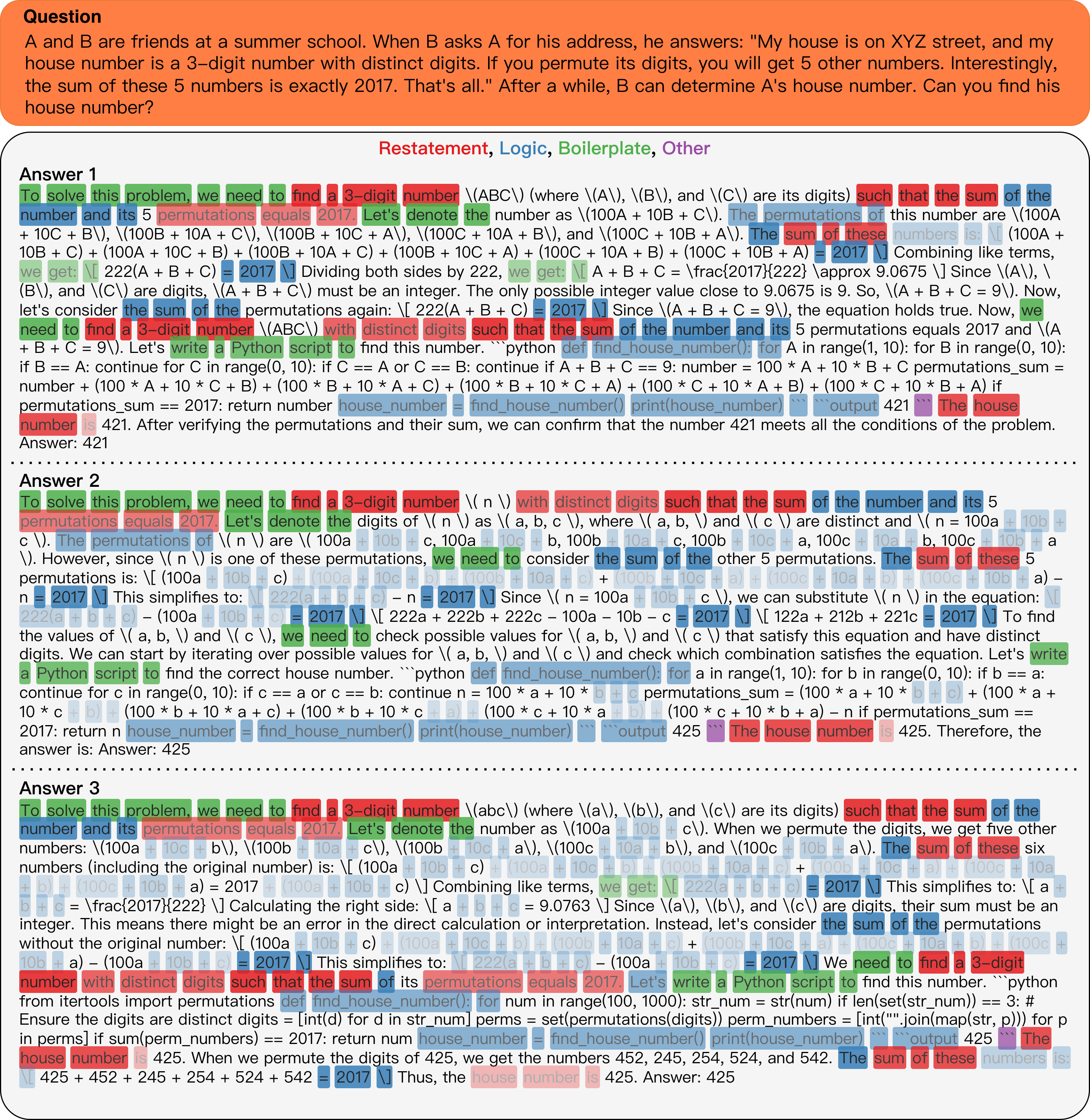}
    \caption{
    Example of repeated $n$-grams extracted from completions of a single prompt.
    High-frequency $n$-grams naturally cluster into four categories: restatement, symbolic or algebraic logic steps, boilerplate phrases, and other tokens. The shading of the $n$-grams indicates their frequency, with darker colors representing higher repetition.
    }
    \label{fig:ngram_case}
\end{figure}

Fig.~\ref{fig:ngram_case} presents the final checkpoint of GRPO training, showing three answers to a single prompt, with n-grams that appear more than 8 times highlighted.

\section{Experimental Details}
\label{app:exp_details}
\subsection{Training Details}
\label{app:mia_training_details}
For the RL-MIA benchmark, we follow ~\citet{tao2025detecting} to train Qwen2.5-7B-Instruct and DeepSeek-Math-
7B-Instruct with the following hyperparameters:
\begin{table}[h]
\centering
\caption{Hyperparameter settings for Qwen2.5-7B-Instruct and Deepseek-math-7b-Instruct in RL-MIA.}
\label{tab:mia_hyperparameters}
\small 
\begin{tabular}{lcc}
\toprule
\textbf{Parameter} & \textbf{Qwen2.5-7B-Instruct} & \textbf{Deepseek-math-7b-Instruct} \\ 
\midrule
Actor learning rate          & $1.0 \times 10^{-6}$ & $1.0 \times 10^{-6}$ \\
Train batch size     & 128            & 128            \\
Max prompt length            & 1024                 & 1024                 \\
Max generation length        & 4096                 & 3072                 \\
Temperature    & 1.0             & 1.0            \\
Samples per prompt ($n$)     & 8                    & 8                    \\
Tensor model parallel (TP)   & 2                    & 2                    \\
Mini batch size           &  2                &  2                \\
Use KL loss                  & No                   & No                   \\
\bottomrule
\end{tabular}
\end{table}
\subsection{Data Details}
\begin{table}[H]
\centering
\caption{RL-MIA data splits for training and evaluation (Selected).}
\label{tab:rl_mia_subset}
\small
\begin{tabular}{llcccc}
\toprule
\textbf{Source} & \textbf{Base RL Corpus (size)} & \textbf{Injected items} & \textbf{Train Size} & \textbf{Occurrences} & \textbf{Detection Tasks} \\ 
\midrule
\textbf{K\&K} & K\&K train: \textbf{950} & K\&K test: \textbf{50} & 950 + 50 & 3 & \textbf{100} \\
\textbf{SAT}  & SAT train: \textbf{450}  & SAT test: \textbf{50}  & 450 + 50 & 3 & \textbf{100} \\ 
\bottomrule
\end{tabular}
\end{table}
\subsection{Implementation Details}
\label{app:implementation_details}
In all experiments, we use vLLM and deploy all models on 8 H800 GPUs with TP=8 or TP=4. During sampling, the default temperature is set to 0.7, and top\_p is set to 0.95. Each model’s output is capped at a max token length of 1024. We found that this token limit provides sufficient detection performance while also speeding up detection time.

\section{Cost Analysis}
\label{app:cost_analysis}
We calculate the average detection time per sample using 32 completions on Open-Reasoner-Zero-7B, as shown in Tab.~\ref{tab:cost_comparison}. While \ours\ requires more time compared to methods like PPL and DIME, with 6.65 seconds per item for 32 completions, the increased time is expected given the need to sample multiple completions for structural diversity analysis. This time is still reasonable and comparable to existing methods, making \ours\ an effective approach for RLVR data exposure detection.

\begin{table}[h]
\centering
\caption{Comparison of average detection latency per sample.}
\label{tab:cost_comparison}
\small
\begin{tabular}{lr}
\toprule
\textbf{Method} & \textbf{Avg. Latency (s/item)} \\ 
\midrule
PPL / Min-K / Min-K++ & 1.43 \\
DIME                  & 2.22 \\
Self-Critique         & 2.67 \\
CDD / Consistency     & 4.29 \\
\midrule
\textbf{\ours\ ($n=16$)} & 4.56 \\
\textbf{\ours\ ($n=32$)} & 6.65 \\
\bottomrule
\end{tabular}
\end{table}

\section{Limitations}
\label{app:limitations}
While \ours\ provides an effective solution for detecting RLVR exposure, there are some limitations. It relies on the assumption that RLVR-induced structural collapse is observable across various models and tasks. Additionally, the method requires generating multiple completions per prompt, which can increase computational cost. Finally, \ours\ focuses on detecting structural collapse but does not directly address broader concerns like potential biases in training data.

\end{document}